\documentclass[11pt,a4paper]{article}
\usepackage{times,latexsym}
\usepackage{url}
\usepackage[T1]{fontenc}

\usepackage{enumitem}
\usepackage{times}

\usepackage{soul}
\usepackage[normalem]{ulem}
\usepackage[hidelinks]{hyperref}
\usepackage[utf8]{inputenc}
\usepackage[small]{caption}
\usepackage{graphicx}
\usepackage{amsmath}
\usepackage{booktabs}
\usepackage{multirow}
\usepackage{amssymb}
\usepackage{footnote}

\usepackage{multirow}
\usepackage[table,xcdraw]{xcolor}

\usepackage{rotating}
\usepackage{makecell}
\usepackage{tcolorbox}

\usepackage{fixmath}
\usepackage{lipsum}                     
\usepackage{xargs}     
\usepackage{subcaption}

\usepackage[symbol]{footmisc}
\usepackage{tablefootnote}

\let\oldbibliography\thebibliography
\renewcommand{\thebibliography}[1]{%
  \oldbibliography{#1}%
  \setlength{\itemsep}{0pt}%
}

\usepackage{bigfoot}

\DeclareNewFootnote{A}

\MakePerPage{footnoteA}
\DeclareNewFootnote{B}

 \usepackage[acceptedWithA]{tacl2018v2}
\usepackage{cleveref}
\crefname{section}{§}{§§}
\Crefname{section}{§}{§§}

\usepackage[acceptedWithA]{tacl2018v2}

\usepackage{xspace,mfirstuc,tabulary}

\newif\iftaclinstructions
\taclinstructionsfalse 
\iftaclinstructions
\renewcommand{\confidential}{}
\renewcommand{\anonsubtext}{(No author info supplied here, for consistency with
TACL-submission anonymization requirements)}
\newcommand{\instr}
\fi

\iftaclpubformat 

\else

\fi

\title{Neuron-level Interpretation of Deep NLP Models: A Survey}

\author{
Hassan Sajjad$^{\clubsuit}$\thanks{\hspace{1.5mm}The authors contributed equally}\hspace{11mm} Nadir Durrani$^{\spadesuit}$\footnotemark[1] \hspace{11mm} Fahim Dalvi$^{\spadesuit}$\footnotemark[1] \\
{\tt hsajjad@dal.ca, \{ndurrani, faimaduddin\}@hbku.edu.qa} \\ 
$^{\clubsuit}$Faculty of Computer Science, Dalhousie University, Canada\thanks{\hspace{1.5mm}The work was done while the author was at QCRI}  \\
$^{\spadesuit}$Qatar Computing Research Institute, HBKU, Doha, Qatar \\ 
}

\date{}

\begin{document}
\maketitle
\begin{abstract}

The proliferation of deep neural networks in various domains has seen an increased need for interpretability of these models. 
Preliminary work done along 
this line and papers that surveyed such, are focused on 
high-level representation analysis. However, a recent branch of work has concentrated on interpretability at a more granular level of analyzing neurons 
within these models. In this paper, we survey the work done on neuron analysis including: i) 
methods to discover and understand neurons in a network, ii) 
evaluation methods, 
iii) major findings including cross architectural comparisons that neuron analysis has unraveled, 
iv) applications of neuron probing such as: controlling the model, domain adaptation etc., and v) a discussion on open issues and future research directions. 
\end{abstract}

\section{Introduction}

Models trained using Deep Neural Networks (DNNs) have constantly pushed the state-of-the-art in 
various Natural Language Processing (NLP) problems, for example Language Modeling \cite{Mikolov:2013:ICLR,devlin-etal-2019-bert} and Machine Translation \cite{SutskeverVL14, bahdanau2014neural} to name a few. Despite this remarkable revolution, the black-box nature of deep neural networks 
has remained major bottleneck in their large scale adaptability -- especially in the applications where fairness, trust, accountability, reliability and ethical decision-making are considered critically important metrics or at least as important as model's performance \cite{lipton2016mythos}. 

This opaqueness of Deep Neural Networks has spurred a new area of research to analyze and understand these models. A plethora of papers have been written in the past five years on interpreting deep NLP models and to answer one question in particular: \textit{\textbf{What knowledge is learned within representations?}} We term 
this 
work as the 
\emph{Representation Analysis}. 

Representation Analysis thrives on post-hoc decomposability, where we analyze 
the embeddings to uncover linguistic (and non-linguistic) concepts\footnote{Please refer to Section~\ref{sec:definitions} for a formal definition.} that are captured as the network is trained towards an NLP task \cite{adi2016fine,belinkov:2017:acl,conneau2018you,liu-etal-2019-linguistic,tenney-etal-2019-bert}.
A majority of the work on \emph{Representation Analysis} has focused on a holistic view of the representations i.e. how much knowledge of a certain concept is learned within representations as a whole (See \newcite{belinkov-etal-2020-analysis} for a survey done on this line of work). Recently, a more fine-grained neuron interpretation has started to gain attention. In addition to 
the holistic view of the representation, \emph{Neuron Analysis} provides insight into a fundamental question: \textbf{\textit{How is knowledge structured within these representations?}} In particular, it targets 
questions such as:

\begin{itemize}
\item What 
concepts are learned within neurons of the network?
\item Are there neurons 
that specialize in learning particular concepts?
\item How localized/distributed and redundantly is the knowledge preserved within neurons of the network?
\end{itemize}

Answers to these questions entail potential benefits beyond understanding the inner workings of models, 
for example: i) controlling bias and manipulating system’s behaviour by identifying relevant neurons with respect to a prediction, ii) model distillation by removing less useful neurons, iii) efficient feature selection by selecting the most salient neurons and removing the redundant ones, iv) neural architecture search by guiding the search with important neurons. 

The work on neuron analysis has explored various 
directions such as: proposing novel methods to 
discover concept neurons 
\cite{Mu-Nips,torroba-hennigen-etal-2020-intrinsic}, analyzing and comparing 
architectures 
using neuron distributions \cite{wu:2020:acl,suau2020finding, durrani-etal-2020-analyzing}, and 
enabling applications of neuron analysis~\cite{bau2018identifying,knowledgeneurons}.
In this survey, we aim to provide a broad perspective of the field with an in-depth coverage of each of these directions. We propose a matrix of seven attributes to compare various neuron analysis methods. Moreover, 
we discuss the open issues and promising future directions in this area.

The survey is organized as follows: 
Section~\ref{sec:definitions} defines the terminologies and formally introduces neuron analysis. Section~\ref{sec:methods} covers various neuron analysis methods and compares them 
using seven attributes.
Section~\ref{sec:evaluation} presents the 
techniques that have been used to evaluate the effectiveness of neuron analysis methods. 
Section~\ref{sec:findings} discusses the findings of neuron analysis methods. 
Lastly 
Section~\ref{sec:applications} showcases various applications of the presented methods and Section~\ref{sec:conclude} 
touches upon the open issues and future research directions. 

\begin{figure*}[]
    \centering
	\includegraphics[width=0.98\linewidth]{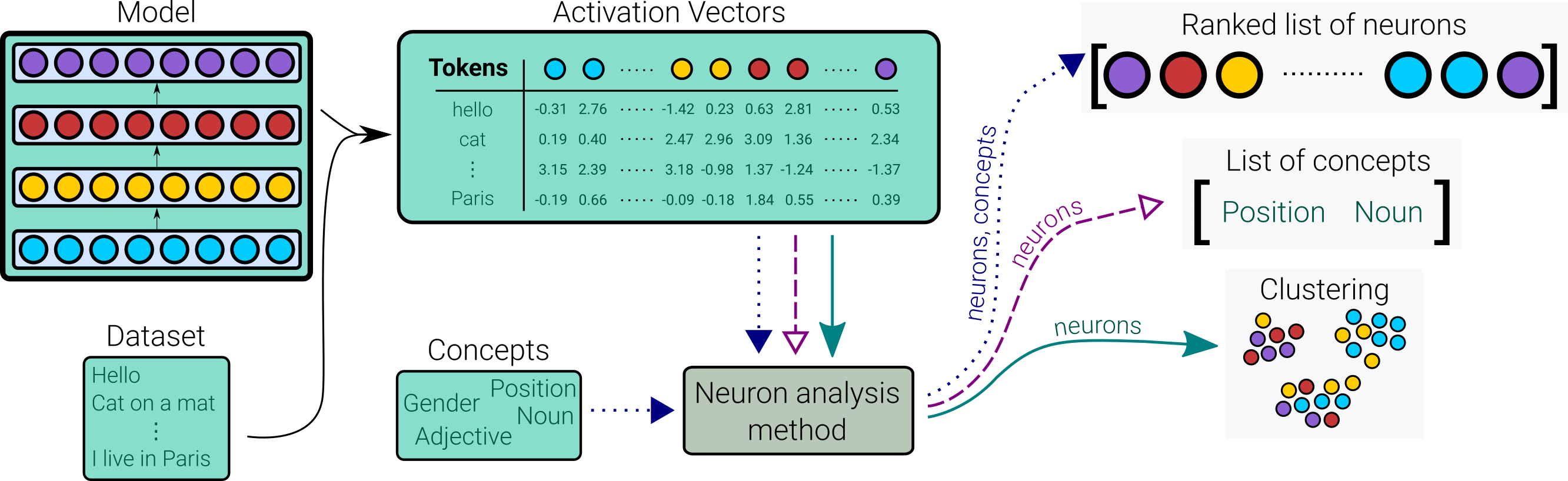}
	\caption{Overview of neuron analysis summarizing the three objectives as discussed in Section~\ref{sec:definitions}}
	\label{fig:overview}
\end{figure*}

\begin{table*}[t]
	\centering
	\footnotesize
	
	\begin{tabular}{c|cccccccc}
	\toprule
		Words & Obama  & receives & Netanyahu  & in  & the & capital & of  & USA \\ 
		\midrule
		Suffix & -- & s & -- & -- & -- & -- & -- & -- \\
		POS  & NNP & VBZ & NNP & IN & DT & NN & IN & NP \\ 
		SEM  & PER & ENS & PER & REL & DEF & REL & REL & GEO \\ 
		Chunk & B-NP & B-VP & B-NP & B-PP & B-NP & I-NP & B-PP & B-NP \\
		CCG & NP & ((S[dcl]$\backslash$NP)/PP)/NP & NP  & PP/NP  & NP/N  & N  & (NP$\backslash$NP)/NP  & NP
        \\
       	\bottomrule
	\end{tabular}
	
    \caption{Example sentences with different word-level concepts. POS: Parts of Speech tags, SEM: Semantic tags, Chunk: Chunking tags, CCG: Combinatory Categorial Grammar tags}
	\label{tab:example-annotation}
\end{table*}

\section{Definitions}
\label{sec:definitions}
In this section, we define the terminologies used in 
the paper and 
the objective of neuron analysis more formally.

\paragraph{Neuron}
Neural networks,
such as RNNs or Transformer models consist of various components such as gates/cells, blocks, layers, attention heads, etc. We use the term \emph{neuron} (also called as \emph{features}, \emph{experts}, and \emph{units} in the literature) to refer to the output of a single dimension from any 
neural network component. 
For example, in the BERT base model, 
the output of a layer block has 768 neurons and the output of an attention head has 64 neurons. 
Moreover, we refer to individual neurons that learn a single concept as \emph{focused neurons}, and a set of neurons that in combination represent a concept as 
\emph{group neurons}. 

\paragraph{Concept}
A concept represents a coherent fragment of knowledge, such as ``a class containing certain objects as elements, where the objects have certain properties'' \cite{stock2010concepts}.
For example, a concept could be lexical: e.g. words ending with suffix ``ed'', morphological: e.g. gerund verbs, or semantic: e.g. names of cities. We loosely define a concept $\mathcal{C}$ as \textbf{\textit{a group of words that are coherent w.r.t to a linguistic property}}. Table \ref{tab:example-annotation} shows an example sentence with different concept annotations.


\paragraph{Objective}
%
%
Figure~\ref{fig:overview} presents an overview of various objectives in neuron analysis. 
Formally, given a model $\mathcal{M}$ and a set of neurons $\mathcal{N}$ (which may consist of all the neurons in the network or a specific subset from particular components like a layer or an attention head) and a concept $\mathcal{C}$, neuron analysis aims to achieve one of the following objectives: 

\begin{itemize}
    
	\item For a concept $\mathcal{C}$, find a ranked list of $|\mathcal{N}|$ neurons
	with respect to the concept (dotted blue line)
	\item Given a neuron $n_i \in \mathcal{N}$, find a set of concepts $|\mathcal{C}|$ the neuron represents (dashed purple line)
	\item Given a set of neurons, find a subset of neurons that encode similar knowledge (solid green line)
\end{itemize}

The former two aim to understand what concepts are encoded within the learned representation. The last objective analyzes how knowledge is distributed across neurons.
 Each neuron $n_i \in \mathcal{N}$ is represented as a vector of activation values over some dataset $\mathcal{D}$. 
Here, every element of the vector corresponds to a word.
For phrase or sentence-level concepts, an aggregation of neuron activations over words in the phrase/sentence is used. Alternatively, \texttt{[CLS]} token representation is also used for transformer models that are transfer learned towards a downstream NLP task. 


\section{Neuron Analysis Methods}
\label{sec:methods}

We have classified the work done on neuron analysis into 5 broader categories of methods, namely: i) visualizations, ii) corpus-based, iii) probing-based, iv) causation-based and v) miscellaneous methods, based on a set of attributes we describe below: 

\begin{itemize}
    \item Scope: Does the method provide global 
    or local 
    interpretation? 
    Global methods accumulate statistics across a set of examples to discover the role of a neuron. 
    Local methods provide interpretation of a neuron in a particular example and may not necessarily reflect its role over a large corpus.
    
    \item Input and Output: What is the input (e.g. a set of neurons or concepts) to the method and what does it output? 
    \item Scalability: Can the method be scaled to a larger set of neurons?
    \item HITL: Does the method require a human-in-the-loop for interpretation?
    \item Supervision: Does the method depend on labeled data to provide interpretation?
    \item Causation: Is the interpretation connected with the model's prediction? 
\end{itemize}

Table~\ref{tab:neuronmethods} summarizes and compares each method in the light of these attributes. We discuss them in detail below.\footnote{Table \ref{tab:appendix:neuronmethods} in Appendix gives a more comprehensive list.}
%

\begin{table*}[t]
\centering
\footnotesize
\begin{tabular}{p{0.19\textwidth}p{0.08\textwidth}p{0.1\textwidth}p{0.1\textwidth}p{0.1\textwidth}p{0.06\textwidth}p{0.1\textwidth}p{0.06\textwidth}}
\toprule
   & Scope & Input & Output  & Scalability & HITL & Supervision & Causation \\
\midrule
\rowcolor[HTML]{D9EAD3} 
\multicolumn{8}{l}{\cellcolor[HTML]{D9EAD3}\textbf{Visualization}}                                                                                                                        \\
\rowcolor[HTML]{D9EAD3} 
\citet{karpathy2015visualizing}  & local        & neuron                 & concept              & low         & yes  & no              & no            \\
\rowcolor[HTML]{A4C2F4} 
\multicolumn{8}{l}{\cellcolor[HTML]{A4C2F4}\textbf{Corpus-based methods}}                                                              
                                     \\
\multicolumn{8}{l}{\cellcolor[HTML]{A4C2F4}{Concept Search}}                                                                                                                  \\
\rowcolor[HTML]{A4C2F4} 
\cellcolor[HTML]{A4C2F4}    \citet{kadar-etal-2017-representation} & global       & neuron                 & concept              & low         & yes  & no              & no            \\
\rowcolor[HTML]{A4C2F4} \citet{Na-ICLR} 
  & global       & neuron                 & concept              & high        & no   & no              & no            \\
\rowcolor[HTML]{A4C2F4} 
\multicolumn{8}{l}{\cellcolor[HTML]{A4C2F4}{Neuron Search}}                                                                                                                  \\
\rowcolor[HTML]{A4C2F4} 
\cellcolor[HTML]{A4C2F4}    \citet{Mu-Nips,suau2020finding,antverg2022on}  & global       & concept                & neurons              & high        & no   & yes             & no            \\
\rowcolor[HTML]{EA9999} 
\multicolumn{8}{l}{\cellcolor[HTML]{EA9999}\textbf{Probing-based methods}}                                                                                                            \\
\rowcolor[HTML]{EA9999} Linear \cite{dalvi:2019:AAAI}     & global       & concept                & neurons              & high        & no   & yes             & no            \\
\rowcolor[HTML]{EA9999} 
Gaussian    \cite{torroba-hennigen-etal-2020-intrinsic}  & global       & concept                & neurons              & high        & no   & yes             & no            \\
\rowcolor[HTML]{D9D9D9} 
\multicolumn{8}{l}{\cellcolor[HTML]{D9D9D9}\textbf{Causation-based methods}}                                                                                                              \\
\rowcolor[HTML]{D9D9D9} 
Ablation    \cite{lakretz-etal-2019-emergence}   & both         & concept/ class         & neurons              & medium      & no   & no              & yes           \\
\rowcolor[HTML]{D9D9D9} 
Knowledge attribution  \cite{knowledgeneurons}   & local        & concept/ class         & neurons              & high        & no   & no              & yes           \\
\rowcolor[HTML]{FFE599} 
\multicolumn{8}{l}{\cellcolor[HTML]{FFE599}\textbf{Miscellaneous methods}}                                                                                                                \\
\rowcolor[HTML]{FFE599} 
Corpus generation \cite{poerner-etal-2018-interpretable}   & global       & neuron                 & concept              & low         & yes  & no              & no            \\
\rowcolor[HTML]{FFE599} 
Matrix factorization \cite{alammar2020explaining}   & local        & neurons                      & neurons      & low         & yes  & no              & no            \\
\rowcolor[HTML]{FFE599} 
Clustering \cite{dalvi-2020-CCFS}      & global        & neurons                      & neurons      & high         & yes  & no    & no            \\
\rowcolor[HTML]{FFE599} 
Multi model search  \cite{bau2018identifying} & global       & neurons                & neurons              & high        & yes  & no              & no     \\ 
\bottomrule
\end{tabular}

\caption{Comparison of neuron analysis methods based on various attributes. The exhaustive list of citations for each method are provided in the text.} 
\label{tab:neuronmethods}
\end{table*}

\subsection{Visualization}
\label{sec:method:visualization}

A simple way to discover the role of a neuron is 
by visualizing its activations and manually identifying 
the underlying concept over a set of sentences~\cite{karpathy2015visualizing,fyshe-etal-2015-compositional,li-etal-2016-visualizing}.
Given that deep NLP models are trained using billions of neurons, it is 
impossible to visualize all 
the neurons. A number of clues have been used to shortlist the 
neurons for visualization, 
for example, selecting saturated neurons, high/low variance neurons, or ignoring dead neurons~\cite{karpathy2015visualizing} when using ReLU activation function.\footnote{Saturated neurons have a gradient value of zero. Dead neurons have an activation value of zero.}

\paragraph{Limitation}

While visualization is a simple approach to find an explanation for a neuron, it has 
some major limitations: i) it is qualitative 
and subjective, ii) it cannot be scaled to the entire network due to an extensive human-in-the-loop effort, iii) it is difficult to interpret polysemous neurons that acquire multiple roles in different contexts, 
iv) 
it is ineffective in identifying \emph{group neurons},
and lastly and v) not all neurons are visually interpretable. Visualization nevertheless remains a useful tool when applied in combination to other interpretation methods that are discussed below.


\subsection{Corpus-based Methods}
\label{sec:method:corpus_selection}

Corpus-based methods discover the role of neurons by aggregating statistics over data activations. 
They establish a connection between a neuron and a concept using co-occurrence between a neuron's activation values and existence of the concept in the underlying input instances (e.g. word, phrases or the entire sentence).
Corpus-based methods are global interpretation methods
as they interpret the role of a neuron over a set of inputs. 
They can be effectively used in combination with the visualization method to reduce the search space 
for finding 
the most relevant portions of data that activates a neuron, thus significantly reducing the human-in-the-loop effort.
Corpus-based methods can be broadly classified into two sets: i) the methods that take a neuron as an input and identify the concept the neuron has learned (\emph{Concept Search}), ii) and others that take a concept as input and identify the neurons learning the concept (\emph{Neuron Search}). 

\paragraph{Concept Search} This set of methods take a neuron as an input and search for a concept that the neuron has learned. They sort the input instances based on the activation values of the given neuron. The top activating instances represent a concept the neuron represents. 
\newcite{kadar-etal-2017-representation} discovered neurons that learn various linguistic concepts using this approach. 
They extracted top-20, 5-gram contexts for each neuron based on the magnitude of activations and manually 
identified the underlying concepts. This manual effort of identifying concepts is cumbersome and requires a 
human-in-the-loop. \newcite{Na-ICLR} addressed this by using lexical concepts of various granularities. Instead of 5-gram contexts, they extracted top-k activating sentences for each neuron. They parsed the 
sentences to 
create concepts (words and phrases) using the nodes of the parse trees.
They then created synthetic sentences that highlight a concept e.g. a particular word occurring in all synthetic sentences. The neurons that activates largely on these sentences are considered to have learned the concept. This methodology is 
useful in analyzing neurons that are responsible for multi-word concepts such as phrases and idiomatic collocations. 
However, the synthetic sentences are often ungrammatical and 
lead towards a risk of identifying neurons that exhibit 
arbitrary behavior (like repetition) instead of concept specific behavior.

\paragraph{Neuron Search} The second class of corpus-based methods 
aim to discover neurons 
for a given concept.
The underlying idea is the same 
i.e. to establish a link between the concept and neurons based on co-occurrences stats, but in the opposite direction. The activation values play a role in weighing these links to obtained a ranked list of neurons against the concept.
\newcite{Mu-Nips} 
achieved this by creating a binary mask of a neuron based on a threshold on its activation values for every sentence in the corpus. Similarly, they created a binary mask for every concept based on its presence or absence in a sentence. They then computed the overlap between a given neuron mask vector and a concept mask vector using intersection-over-union (IoU),
 and use these to generate compositional explanations.  Differently from them, \newcite{suau2020finding} used the values of neuron activations as prediction scores and computed the average precision per neuron and per concept. Finally, \newcite{antverg2022on} considered the mean activation values of a neuron with respect to instances that posses the concept of interest.

The two methods give an alternative view to neuron interpretation. While \emph{Neuron Search} methods aim to find the neuron that has learned a concept, \emph{Concept Search} methods generate explanations for neurons by aligning them with a concept.


\paragraph{Limitation}

The corpus-based methods do not model the selection of \emph{group neurons} that work together to learn a concept. Concept Search methods consider every neuron independently. Similarly, Neuron Search methods do not find the correlation of a group of neurons with respect to the given concept.

\subsection{Probing-based Methods}
\label{sec:method:probing_classifier}


Probing-based methods train 
diagnostic classifiers~\cite{hupkes2018visualisation} over activations to identify neurons with respect to pre-defined concepts. 
They are a global interpretation methods that discover 
a set of neurons with respect to each concept using supervised data annotations. 
They are highly scalable, and can be easily applied on a large set of neurons and 
over a large set of concepts. 
%
In the following, we cover two types of classifiers used for probing.

\paragraph{Linear Classifiers}

The idea 
is to train a linear classifier towards the concept of interest, using the activation vectors generated by the model being analyzed. 
The weights assigned to neurons (features to the classifier) serve as their importance score with respect to the concept. The regularization of the classifier directly effects the weights and therefore the ranking of neurons. \newcite{Radford} used L1 regularization which forces the classifier to learn spiky weights, indicating the selection of very few specialized neurons learning a concept, while setting the majority of neurons' weights to zero. \newcite{lakretz-etal-2019-emergence} on the other hand used L2 regularization to encourage grouping of features. This translates to discovering \emph{group neurons} that are jointly responsible for a concept. \newcite{dalvi:2019:AAAI} used ElasticNet regularization which combines the benefits of L1 and L2, accounting for both highly correlated 
\emph{group neurons} and 
specific \emph{focused neurons} with respect to a concept.

\paragraph{Limitation} A pitfall to 
probing classifiers is whether a probe 
faithfully reflects the concept 
learned within the representation or just 
memorizes the task~\cite{hewitt-liang-2019-designing,zhang-bowman-2018-language}. Researchers have mitigated this pitfall for some analyses by using random initialization of neurons~\cite{dalvi:2019:AAAI} and control tasks~\cite{durrani-etal-2020-analyzing} to demonstrate that the knowledge is possessed within the neurons and not due to the probe's capacity for memorization. Another discrepancy in the neuron probing framework, that especially affects the linear classifiers, is that variance patterns in neurons differ strikingly across the layers. \newcite{sajjad2021:znorm} suggested to apply z-normalization as a pre-processing step to any neuron probing method to alleviate this issue.



\paragraph{Gaussian Classifier}
\newcite{torroba-hennigen-etal-2020-intrinsic} trained a generative classifier with the assumption that neurons exhibit a Gaussian distribution. They fit a multivariate Gaussian over all neurons and 
extracted individual probes for single neurons. A 
caveat to their approach is that activations do not always follow a \emph{Gaussian prior} in practice -- hence restricting their analysis to only the neurons that satisfy this criteria. Moreover, the interpretation is limited to single neurons and identifying groups of neurons requires an expensive greedy search. 

%

\paragraph{Limitation}
In addition to the shortcomings discussed above, a major limitation of probing-based methods is the requirement of supervised data for training the classifier, thus limiting the analysis only to pre-defined or annotated concepts.

\subsection{Causation-based methods}
\label{sec:method:causal}

The methods we have discussed so far are limited to identifying neurons that have learned the encoded concepts. 
They do not inherently reflect 
their importance towards the model's performance. 
Causation-based methods 
identify neurons with respect to model's prediction.

\paragraph{Ablation}
The central idea behind ablation is to notice the affect of a neuron on model's performance by varying its value. This is done either by clamping its value to zero or a fixed value 
and observe
the change in network's performance.
Ablation 
has been effectively used to find i) salient neurons with respect to a model (unsupervised), ii) salient neurons with respect to a 
particular output class in the network (supervised). The former identifies neurons that incur a large drop in model's performance 
when ablated~\cite{li-etal-2016-visualizing}. The latter selects neurons that cause the model to flip its prediction with respect to a certain class~\cite{lakretz-etal-2019-emergence}. Here, the output class serves as the concept against which we want to find the salient neurons. 

\paragraph{Limitation}  
Identifying 
\emph{group neurons} require ablating all possible combinations of neurons which is an NP-hard problem~\cite{binshtok:2007:AAAI}. Several researchers have tried to circumvent this by using 
leave-one-out estimates~\cite{ZintgrafCAW17}, beam search~\cite{feng-etal-2018-pathologies}, by learning end-to-end differentiable prediction model~\cite{de-cao-etal-2020-decisions} and by using correlation clustering to group similar neurons before ablation~\cite{dalvi-2020-CCFS}. Nevertheless all these approaches are approximations and may incur search errors.

\paragraph{Knowledge Attribution Method}

Attribution-based methods highlight the importance of input features and neurons with respect to a  prediction~\cite{DBLP:journals/corr/abs-1805-12233,shappely_NIPS2017_7062,tran2018importance}. \newcite{knowledgeneurons} 
used an attribution-based method to identify salient neurons with respect to a relational fact. They hypothesized that factual knowledge is stored in the neurons of the feed-forward neural networks of the Transformer model and used integrated gradient~\cite{ig} to identify top neurons that express a relational fact. 
The work of \newcite{knowledgeneurons} shows the applicability of attribution methods in discovering causal neurons with respect to a concept of interest and is a promising research direction.

\paragraph{Limitation}
The attribution-based methods highlight salient neurons with respect to a prediction. What concepts these salient neurons have learned is unknown. \newcite{knowledgeneurons} worked around 
this by limiting their study to model classes where each class serves as a concept. Attribution-based methods can be enriched by
complementing them with other neuron analysis methods such as corpus search that associate salient neurons to a concept. 
 
\subsection{Miscellaneous Methods}
\label{sec:method:unsupervised}

In this section, we cover a diverse set of 
methods that do not fit in the above defined categories. 

\paragraph{Corpus Generation}
A large body of neuron analysis methods identify neurons with respect to pre-defined concepts and the scope of search is only limited to the corpus used to extract the activations. 
It is possible that a neuron represents a diverse concept which is not featured in the corpus. The \emph{Corpus Generation} method addresses this problem by generating novel sentences that maximize a neuron's activations. These sentences unravel hidden information about a neuron, facilitating the annotator to better describe its role. Corpus generation has been widely explored in Computer Vision. For example, \newcite{erhan_gradient_ascent_techreport} used gradient ascent to generate synthetic input images that maximize the activations of a neuron. However, a 
gradient ascent can not be directly applied in NLP, because 
of the discrete inputs. \newcite{poerner-etal-2018-interpretable} worked around this problem by using \emph{Gumble Softmax} 
and showed their method to surpass Concept Search method
\cite{kadar-etal-2017-representation} in interpreting neurons. 

\paragraph{Limitation} Although the corpus generation method has the benefit of generating novel patterns that explain a neuron beyond the space of the underlying corpus, it often generates nonsensical patterns and sentences that are difficult to analyze in isolation. A thorough 
evaluation is necessary to know its true potential and efficacy in NLP. 

\paragraph{Matrix Factorization}

Matrix Factorization (MF) method decomposes a large matrix into a product of smaller matrices of factors, where each factor represents a group of elements performing a similar function. Given a model, the activations of an input sentence form a matrix. MF can be effectively applied to decompose the activation matrix into smaller matrices of factors where each factor consists of a set of neurons that learn a concept. MF is a local interpretation method. It is commonly used in analyzing vision models~\cite{olah2018the}. We could not find any research using MF on the NLP models. To the best of our knowledge, \newcite{alammar2020explaining} is the only blog post that introduced them 
in the NLP domain. 

\paragraph{Limitation} Compared to the previously discussed unsupervised methods, MF has an innate benefit of 
discovering \emph{group neurons}. However, it is still non-trivial to identify the number of groups (factors) to decompose the activations matrix into. Moreover, the scope of the method is limited to local interpretation.

\paragraph{Clustering Methods}
Clustering is another effective way to analyze groups of neurons in an unsupervised fashion. The intuition is that if a group of neurons learns a specific concept, then their activations would 
form a cluster. \newcite{mayes_under_hood:2020} used UMAP~\cite{mcinnes2020umap} to project 
activations to a low dimensional space and performed K-means clustering to group neurons. \newcite{dalvi-2020-CCFS} aimed at identifying redundant neurons in the network. They first computed correlation between neuron activation pairs and used hierarchical clustering to group them. The neurons with highly correlated behavior are clustered together and are considered redundant in the network. 

\paragraph{Limitation} Similar to the Matrix Factorization method, the number of clusters is a hyperparameter that needs to be pre-defined or selected empirically. A small number of clusters may result in dissimilar neurons in the same group while a large number of clusters may lead to similar neurons split in different groups. 

\paragraph{Multi-model Search}
Multi-model search is based on the intuition that salient information is shared across the models trained towards a task i.e. if a concept is important for a task then all models optimized for the task should learn it. The search involves identifying neurons that behave similarly across the models.  \newcite{bau2018identifying} used Pearson correlation to compute a similarity score of each neuron of a model with respect to the neurons of other models. They aggregated the correlations for each neuron using several methods with the aim of highlighting different aspects of the model. 
More specifically, they used \emph{Max Correlation} to capture concepts that emerge strongly in multiple models, \emph{Min Correlation} to select neurons that are correlated with many models though they are not among the top correlated neurons, \emph{Regression Ranking} to find individual neurons whose information is distributed among multiple neurons of other models, and \emph{SVCCA}~\cite{NIPS2017_7188_svcca} to capture information that may be distributed in fewer dimensions than the whole representation.

\paragraph{Limitation} All the methods discussed in this section, require human-in-the-loop to provide explanation for the underlying neurons. They can nevertheless be useful in tandem with the other interpretation methods. For example \newcite{dalvi:2019:AAAI} intersected the neurons discovered via the probing classifier and the multi-model search to describe salient neurons in the NMT models.

\section{Evaluation}
\label{sec:evaluation}


In this section, we survey the evaluation methods used to measure the correctness of the neuron analysis methods. 
Due to the absence of interpretation benchmarks, it is difficult to precisely define ``correctness''. Evaluation 
methods in interpretation mostly resonate with the underlying method to discovered salient neurons. For example visualization methods often require qualitative evaluation via human in the loop, probing methods claim correctness of their rankings using classifier accuracy as a proxy. 
\newcite{antverg2022on} highlighted 
this discrepancy 
and suggested to disentangle the analysis methodology from the evaluation framework, for example by using a principally different evaluation method compared to the underlying neuron analysis method.
In the following, we summarize various evaluation methods and their usage in the literature.

\subsection{Ablation}

While ablation has been used to discover salient neurons for the model, it has also been used to evaluate the efficacy of the selected neurons. More concretely, given a ranked list of neurons (e.g. the output of the probing method), we ablate neurons in the model in the order of their importance and measure the effect on the performance. 
%
The idea is that removing the top neurons should result in a larger drop in performance compared to randomly selected neurons.
\newcite{dalvi:2019:AAAI, durrani-etal-2020-analyzing} 
used ablation in the probing classifier to demonstrate correctness of their neuron ranking method. 
Similarly \newcite{bau2018identifying} showed that ablating the most salient neurons, discovered using multi-model search, in NMT models lead to a much bigger drop in performance as opposed to removing randomly selected neurons.




\subsection{Classification Performance}

Given salient neurons with respect to a concept, a simple method to evaluate their correctness is to train a classifier using them as features and predict the concept of interest. The performance of the classifier relative to a classifier trained using random neurons and least important neurons is used as a metric to gauge the efficacy of the selected salient neurons. However, it is important to ensure that the probe is truly representing the concepts encoded within the learned representations and not memorizing them during classifier training. \newcite{hewitt-liang-2019-designing} introduced Controlled Tasks Selectivity as a measure to gauge this.
\newcite{durrani-etal-2020-analyzing} adapted controlled tasks for neuron-probing to show that their probes indeed reflect the underlying linguistic tasks.




\subsection{Information Theoretic Metric}

Information theoretic metrics such as mutual information have also been used to interpret representations of deep NLP models~\cite{voita2020informationtheoretic,pimentel2020informationtheoretic}. Here, the goal is to measure the amount of information a representation provides about a linguistic properties. \newcite{torroba-hennigen-etal-2020-intrinsic} used mutual information to evaluate the effectiveness of their Gaussian-based method 
by calculating
the mutual information between subset of neurons and linguistic concepts.

\subsection{Concept Selectivity}

Another evaluation method derived from \emph{Concept Search} methodology measures the alignment between neurons and the discovered concept, by weighing how selectively each neuron responds to the concept \cite{Na-ICLR}. Selectivity is computed by taking a difference between average activation value of a neuron over a set of sentences where the underlying concept occurs and where it doesn't. A high selectivity value is obtained when a neuron is sensitive to the underlying concept and not to other concepts.

\subsection{Qualitative Evaluation}

\emph{Visualization} has been used as a qualitative measure to evaluate the 
selected neurons. For example, \newcite{dalvi:2019:AAAI} visualized the top neurons and showed that they focus on very specific linguistic properties. They also visualized top-k activating words for the top neurons per concept to demonstrate the efficacy of their method. Visualization can be a very effective tool to evaluate the interpretations when it works in tandem with other methods e.g. using Concept Search or Probing-based methods to reduce the search space towards only highly activating concepts or the most salient neurons for these concepts respectively.


\section{Findings}
\label{sec:findings}
Work done on neuron interpretation in NLP is predominantly focused on the questions such as: \textit{i) what concepts are learned within neurons? ii) 
how the knowledge is structured within representations?} We 
now iterate through various 
findings that 
the above described neuron analysis methods unravelled. Based on our main driving questions, we classify 
these into two broad categories: \textit{i) concept discovery and ii) architectural analysis}. 

\subsection{Concept Discovery}
\label{sec:cd}

In the following, we survey what lexical concepts or core-linguistic phenomenon 
are learned by the neurons in the network. 

\subsubsection{Lexical Concepts} 

Some of the research done on neuron analysis particularly the work using visualization and concept search methods identified neurons that capture lexical concepts. 



\paragraph{Visualizations} \newcite{karpathy2015visualizing}  found neurons that learn position of a word in the input sentence: 
activating positively in the beginning, becoming neutral in the middle and negatively towards the end. \newcite{li-etal-2016-visualizing} found intensification neurons that activate for words that intensify a sentiment. For example ``I like this movie \textbf{a lot}'' or ``the movie is \textbf{incredibly} good''. Similarly they discovered neurons that captured ``negation''.  Both intensification neurons and sentiment neurons are relevant for the sentiment classification task, for which the understudied model was trained. 

\paragraph{Concept Search} \newcite{kadar-etal-2017-representation} identified neurons that capture related groups of 
concepts in a multi-modal image captioning task.  For example, they discovered neurons that learn electronic items ``camera, laptop, cables'' and salad items ``broccoli, noodles, carrots etc''. Similarly \newcite{Na-ICLR} found neurons that learn lexical concepts related to legislative terms, e.g. ``law, legal'' etc. They also found neurons that 
learn phrasal concepts. \newcite{poerner-etal-2018-interpretable} showed that \emph{Concept Search} can be enhanced via \emph{Corpus Generation}. They provided finer interpretation of the neurons by generating synthetic instances. For example, they showed that a ``horse racing'' neuron identified via concept search method was in fact a  general ``racing'' neuron by generating novel contexts against this neuron.



\subsubsection{Linguistic Concepts}

A number of studies 
probed for neurons that capture core-linguistic concepts such as 
morphology,
semantic tags, etc. Probing for linguistic structure is important to understand models' capacity to generalize~\cite{MarasovicGradient2018NLP}.\footnote{but is not the only reason to carry such an analysis.}
For example, the holy grail in machine translation is that a proficient model needs to be aware of word morphology, grammatical structure, and semantics to do well~\cite{Vauquois68, jones-etal-2012-semantics}. Below we discuss major findings along this line of work:


\textbf{Neurons specialize in core linguistic concepts}
\newcite{dalvi:2019:AAAI} 
in their analysis of LSTM-based NMT models found neurons that capture 
core linguistic concepts such as nouns, verb forms, numbers, articles, etc. 
They also showed that \textbf{the number of neurons responsible for a concept varies based on the nature of the concept.} 
For example: closed class\footnote{closed class concepts are part of language where new words are not added as the language evolves, for example functional words such as \emph{can, be} etc. In contrast open class concepts are a pool where new words are constantly added as the language evolve, for example "chillax" a verb formed blending "chill" and "relax".} concepts such as \emph{Articles} (morphological category), \emph{Months of Year} (semantic category) are localized to fewer neurons, whereas open class concepts such as \emph{nouns} (morphological category) or \emph{event} (semantic category) are distributed among a large number of neurons. 
%

\textbf{Neurons exhibit monosemous and polysemous behavior.} \newcite{xin-etal-2019-part} found neurons exhibiting a variety of roles where a few neurons were 
exclusive to a single concept while 
others were polysemous in nature and captured several concepts. \newcite{suau2020finding} discovered neurons that capture different senses of a word. Similarly, \newcite{bau2018identifying} found a switch neuron that activates positively for present-tense verbs and negatively for the past tense verbs.

\textbf{Neurons capture syntactic concepts and complex semantic concepts.} 
\newcite{lakretz-etal-2019-emergence} discovered neurons that capture subject-verb agreement within LSTM gates. \newcite{karpathy2015visualizing} also found neurons that activate within quotes and brackets capturing long-range dependency. \newcite{Na-ICLR} aligned neurons with syntactic parses to show that neurons learn syntactic phrases.
%
\newcite{esther_causativity_iwcs21} analyzed complex semantic properties underlying a given sentence. 

\subsubsection{Salient Neurons for Models} 

In contrast to analyzing neurons with respect to a pre-defined concept, researchers also interpreted
the concepts 
captured in the most salient neurons of the network. For example, in the analysis of the encoder of 
LSTM-based models, \newcite{bau2018identifying} used Pearson correlation to discover salient neurons in the network. They
found neurons that learn position of a word in the sentence
among the most important neurons. Other 
neurons found 
included parentheses, punctuation and conjunction neurons. 
Moreover, \newcite{LiMJ16a} found that the two most salient neurons in 
\emph{Glove} 
were the frequency neurons that play an important role in all predictions. 

The question of whether core-linguistic concepts are important for the end performance 
has been a less explored area. \newcite{dalvi:2019:AAAI} compared neurons learning morphological concepts and semantic concepts with unsupervised ranking of neurons with respect to their effect on the end performance. They found that the \textbf{model is more sensitive to the top neurons obtained using unsupervised ranking compared to linguistic concepts}. They showed that the unsupervised ranking of neurons is dominated by position information and other closed class categories such as conjunction and punctuation which according to the ablation experiment are more critical concepts for the end performance than linguistic concepts. 

\subsection{Architectural Analysis}
Alongside studying what concepts are captured within deep NLP models, researchers have also studied: i) how these concepts are organized in the network? ii) how distributed and redundant they are? and iii) how this compares across 
architectures? Such an analysis is helpful in better understanding of the network and can be potentially useful in architectural search and model distillation.

\subsubsection{Information Distribution} Human languages are hierarchical in structure where morphology and phonology sit at the bottom followed by lexemes, followed by syntactic structures. 
Concepts such as semantics and pragmatics are placed on the top of the hierarchy. \newcite{durrani-etal-2020-analyzing} analyzed linguistic hierarchy by studying the spread of neurons across layers in various pre-trained language models. They extracted salient neurons with respect to different linguistic concepts (e.g. morphology and syntax) and found that \textbf{neurons that capture word morphology were predominantly found in the lower and middle layers and those learning about syntax were found at the higher layers.} The observation was found to be true in both LSTM- and the transformer-based architectures, and are inline with the findings of representation analysis~\cite{liu-etal-2019-linguistic,tenney-etal-2019-bert,belinkov-etal-2020-linguistic}.
Similarly \newcite{suau2020finding} analyzed sub-modules within GPT and RoBERTa transformer blocks and showed that lower layers within a transformer block accumulate more salient neurons than higher layers on the tasks of word sense disambiguation or homograph detection. They also found that the neurons that learn homographs are distributed 
across the network as opposed to sense neurons that were more predominantly found at the lower layers. 

\subsubsection{Distributivity and Redundancy} While it is exciting to see that networks somewhat preserve linguistic hierarchy, 
many authors found that information is not discretely preserved at any individual layer, but is distributed 
and is redundantly present in the network. This is an artifact of various training choices such as dropout that encourages the model to distribute knowledge across the network.
For example, \newcite{LiMJ16a} found specialized frequency neurons in a \emph{GloVe} model trained without dropout, as opposed to the variant trained with dropout where the information was more redundantly available.
%
\newcite{dalvi-2020-CCFS} showed that a significant amount of redundancy existed within pre-trained models. They showed that 85\% of the neurons across the network are redundant and at least 92\% of the neurons can be removed when optimizing towards a downstream task in feature-based transfer learning.

\begin{figure*}[]
    \centering
    \begin{subfigure}[b]{0.23\linewidth}
    \centering
    \includegraphics[width=\linewidth]{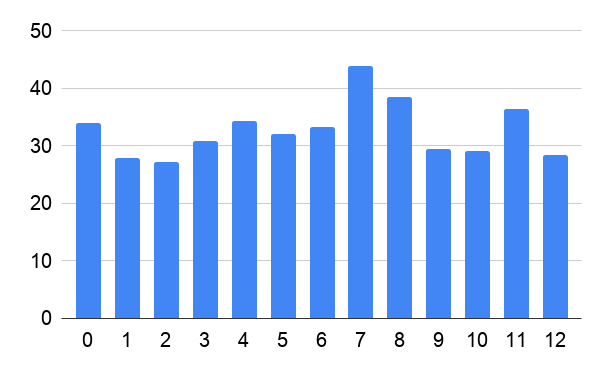}
    \caption{POS -- BERT}
    \label{fig:posbert}
    \end{subfigure}
    \begin{subfigure}[b]{0.23\linewidth}
    \centering
    \includegraphics[width=\linewidth]{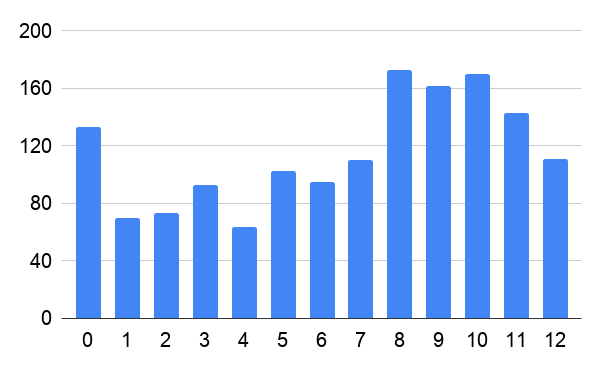}
    \caption{CCG -- BERT}
    \label{fig:ccgbert}
    \end{subfigure}
    \begin{subfigure}[b]{0.23\linewidth}
    \centering
    \includegraphics[width=\linewidth]{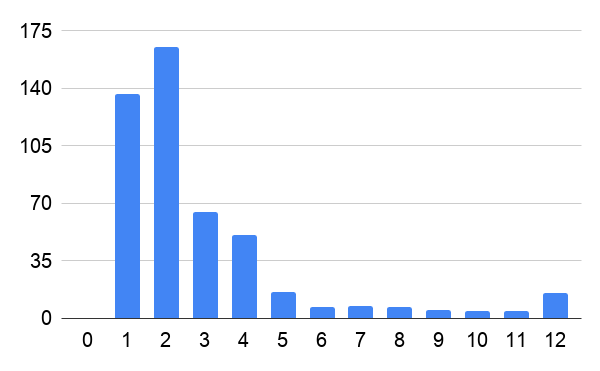}
    \caption{POS -- XLNet}
    \label{fig:posxlnet}
    \end{subfigure}    
    \begin{subfigure}[b]{0.23\linewidth}
    \centering
    \includegraphics[width=\linewidth]{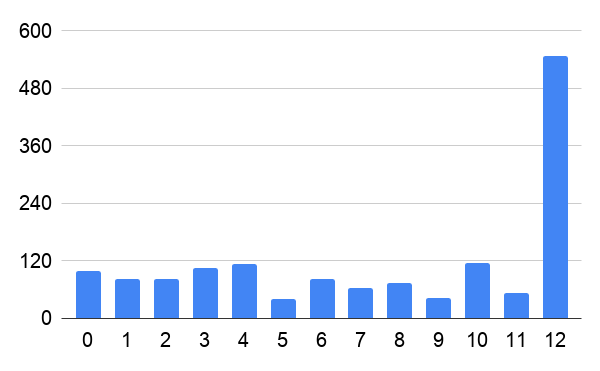}
    \caption{CCG -- XLNet}
    \label{fig:ccgxlnet}
    \end{subfigure}    
    
    \caption{Distribution of top neurons spread across different layers for each task. X-axis = Layer number, Y-axis = Number of neurons selected from that layer -- Figure borrowed from ~\citet{durrani-etal-2020-analyzing}}
    \label{fig:layerwise}
\end{figure*}

\subsubsection{Comparing Architectures} The distribution of neurons across the network has led researchers to draw interesting cross-architectural comparisons. 
\newcite{wu:2020:acl} performed correlation clustering of neurons across 
architectures 
and found that different architectures may have similar representations, but their individual neurons behave differently. 
\newcite{torroba-hennigen-etal-2020-intrinsic} compared neurons in contextualized (BERT) embedding with neurons in the static embedding (fastText) and found that fastText required two neurons to capture any morphosyntactic phenomenon as opposed to BERT which required up to 35 neurons to obtain the same performance. \newcite{durrani-etal-2020-analyzing} showed that the \textbf{linguistic knowledge in BERT (auto-encoder) is highly distributed across the network as opposed to XLNet (auto-regressive) where neurons from a few layers are mainly responsible for a concept} (see Figure~\ref{fig:layerwise}). Similarly \newcite{suau2020finding} compared RoBERTa and GPT (auto-encoder vs. generative) models and found differences in the distribution of expert neurons. \newcite{durrani-FT} extended the cross-architectural comparison towards fine-tuned models. They showed that after fine-tuning on GLUE tasks, the neurons capturing linguistic knowledge are regressed to lower layers in RoBERTa and XLNet as opposed to BERT where it is still retained at the higher layers. 

\subsection{Summary of Findings}
\label{sec:findingSummary}

Below is a summary of the key findings that emerged from the work we covered in this survey. Neurons learned within Deep NLP models capture
non-trivial linguistic knowledge ranging from 
lexical phenomenon such as morphemes, words and multi-word expressions to highly complex global phenomenon such as  semantic roles and syntactic dependencies. Neuron analysis resonates with the findings of representation analysis \cite{belinkov:2017:acl,belinkov-etal-2017-evaluating,tenney-etal-2019-bert, liu-etal-2019-linguistic} in demonstrating that the networks follow linguistic hierarchy. Linguistic neurons are distributed across the network based on their complexity, with lower layers focused on the lexical concepts and middle and higher layers learning global phenomenon based on long-range contextual dependencies. While the networks preserve linguistic hierarchy, many authors showed that information is not discretely preserved, but is rather distributed and redundantly present in the network. It was also shown that a small optimal subset of neurons w.r.t any concept can be extracted from a network. On another dimension, a few works showed that some concepts are localized to fewer neurons while others are distributed to a large group. Finally, some interesting cross architectural analyses were drawn based on how the neurons are distributed within their layers.


\section{Applications}
\label{sec:applications}

Neuron analysis leads to various applications beyond interpretation of deep models. In this section, we present several applications of neuron analysis: i) controlling model's behavior, ii) model distillation and efficiency, iii) domain adaptation, and iv) generating compositional explanations.

\subsection{Controlling Model's Behavior}
\label{sec:manipulation}

Once we have identified neurons that capture a certain concept learned in a model,  
these can 
be utilized for controlling the model's behavior w.r.t to that concept. 
\newcite{bau2018identifying} identified \emph{Switch Neurons} in 
NMT models that activate positively for the present-tense verbs and negatively for the past-tense verbs. By manipulating the values of 
these neurons, 
they were able to successfully change output translations from present to past tense during inference. The authors additionally found neurons that capture \emph{gender} and \emph{number agreement} concepts and manipulated them to control the system's output. Another effort along this line was carried by \newcite{suau2020finding}, where they 
manipulated the neurons responsible for a concept in the GPT model and generated sentences around specific topics of interest. Recently~\newcite{knowledgeneurons} manipulated salient neurons of relational facts and demonstrated  their ability to update and erase knowledge about a particular fact.
Controlling 
model's behavior using neurons 
enables on-the-fly manipulation of output, 
for example it can be used to debias the output of the model against sensitive attributes like race and gender. 




\subsection{Model Distillation and Efficiency}
Deep NLP models are trained using hundreds of millions of parameters, limiting their applicability in computationally constrained environments. Identifying salient neurons and sub-networks can be useful for model distillation and efficiency. \newcite{dalvi-2020-CCFS} devised an efficient feature-based transfer learning procedure, stemmed from their redundancy analysis. By exploiting layer and neuron-specific redundancy in the transformer models, they were able to reduce the feature set size to less than 10\% neurons for several tasks while maintaining more than 97\% of the performance. The procedure achieved a speedup of up to 6.2x in computation time for sequence labeling tasks as opposed to using all the features.

\subsection{Domain Adaptation}

Identifying the salient neurons with respect to a domain can be effectively used for domain adaptation and generalization. \newcite{gu2021pruningthenexpanding} proposed a domain adaptation method using neuron pruning to target the problem of catastrophic forgetting of the general domain when fine-tuning a model for a target domain. They introduced a three step adaptation process:  
i) rank neurons based on their importance, ii) prune the unimportant neurons from the network and retrain with student-teacher framework, iii) expand the network to its original size and fine-tune towards in-domain, freezing the salient neurons and adjusting only the unimportant neurons. Using this approach helps 
in avoiding catastrophic forgetting of the general domain while also obtaining optimal performance on the in-domain data. 


\subsection{Compositional Explanations}
\begin{figure}[]
    \centering
    \includegraphics[width=0.95\linewidth]{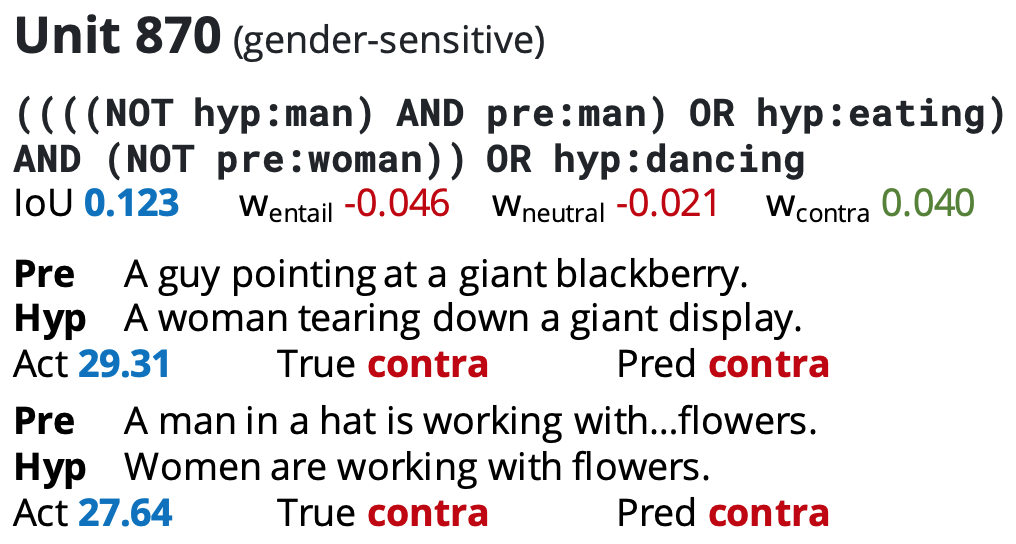}
    \caption{Compositional explanation using neuron 870 on the NLI task -- Figure borrowed from \newcite{Mu-Nips}}
    \label{fig:composition}
\end{figure}

Knowing the association of a neuron with a concept enables explanation of model's output. \newcite{Mu-Nips} identified neurons that learn certain concepts in vision and NLP models. Using a composition of logical operators, they provided an explanation of model's prediction. 
Figure~\ref{fig:composition} presents an explanation using a gender-sensitive neuron. The neuron activates for contradiction when the premise contains the word \textit{man}. Such explanations provide a way to generate adversarial examples that change model's predictions. 

\section{Open Issues and Future Directions}
\label{sec:conclude}

In the following section, we discuss several open issues and limitations related to methods, evaluation and datasets. Moreover, we provide potential future directions vital to the progress of neuron and model interpretation.

\begin{itemize}
    \item DNNs are distributed in nature which encourages groups of neurons to work together to learn a concept. The current analysis methods, at large, ignore interaction between neurons while discovering neurons with respect to a concept. Trying all possible combination of neurons is a computationally intractable problem. A linear classifier using ElasticNet regularization~\cite{dalvi:2019:AAAI} considers grouping of features during training -- however, it's effectiveness in handling grouped neurons has not been empirically validated. Evolutionary algorithms\footnote{\url{https://en.wikipedia.org/wiki/Evolutionary_algorithm}} do not make any assumption of the underline distribution of the features and they have been effectively used for feature selection of multivariate features. Exploring them for neuron selection is a promising research direction to probe towards latent concepts in these models.
    
    \item A large number of interpretation studies rely on human-defined linguistic concepts to probe a model. It is possible that the models do not strictly adhere to the human-defined concepts and learn novel concepts about the language. 
    This results in an incorrect or incomplete analysis. 
    Several researchers \cite{michael-etal-2020-asking, dalvi2022discovering,sajjad-etal-2022-analyzing} made strides in this direction by analyzing hidden structures in the input representations in an unsupervised manner. They discovered existence of novel structures not captured in the human defined categories. 
    \newcite{dalvi2022discovering} also proposed
    BERT ConceptNet, 
    a manual annotation of the latent concepts in BERT. Introducing similar datasets across other models 
    enables model-centric interpretation, and is a promising research direction. 
    
    \item While a lot of work has been done on analyzing how knowledge is encoded within the learned representations, the question whether it is used by the model during prediction is a less explored area \cite{feder-etal-2021-causalm, elazar-etal-2021-amnesic}. 
    Ablation and knowledge attribution methods are two neuron interpretation methods that intrinsically use causal relation to select concept neurons. A few other studies evaluated the causal relation of the selected concept neurons via ablation or by clamping their activation values~\cite{bau2018identifying, suau2020finding} and observed the change in model's prediction. However, most of the studies do not take into account the causal relation as part of the method or the evaluation of their method. The causal relation with respect to concept neurons is important to understand their importance to overall prediction and it leads way towards practical applications such as debiasing, model distillation and domain adaptation.

    \item The work on neuron interpretation lacks standard evaluation benchmarks, and therefore studies conducted on identical models are not comparable. For example, there exists no gold annotation of neurons with respect to a certain dataset or a class. 
    The curation of standard evaluation benchmarks 
    is an essential step towards improving methods of interpretation of deep neural network models.
    
    \item The neuron analysis methods vary in their theoretical foundations as well as the perspective they aim to capture with respect to a given concept. This results in a selection of neurons that may not strictly align across all methods.
    %
    For example, \emph{Visualization}, \emph{Neuron Search} and \emph{Corpus Search} discover neurons that are highly focused on a specific task (like "less" suffix or POS "TO" concepts), while \emph{Probing-based} methods discover ranking of neurons that highlight grouping behavior within the neurons targeting broad concepts like POS "Nouns". 
    Therefore, the choice of which neuron interpretation method to use is not straightforward and depends on various factors such as the nature of the concept to investigate, the availability of supervised data for the concept of interest etc. Apart from these high-level guiding principles, a thorough comparison of methods with respect to the nature of the concept of interest is needed to fully understand the strengths and weaknesses of each approach. \newcite{antverg2022on} is one such effort in this direction that compares three neuron interpretation methods. 
        
    \item Neuron-level interpretation opens door for a number of applications useful for the successful deployment of DNN systems (Section \ref{sec:applications}). However, most of the research conducted in this direction is preliminary. For example, there are many open research questions in \textbf{controlling system's behaviour} using neurons such as: i) are all concepts manipulatable? ii) how to identify neurons that can be controlled to change the output? iii) is high distributiveness a hindrance for controlling model's behavior? iv) and whether disentangled ~\cite{bengio_disentangled} and sparse models~\cite{frankle2018_lotterticket} may serve as a better alternate on this front? Addressing these questions will enable a more reliable control of the deep NLP models and entail numerous applications such as removing bias and adapting the system to novel domains.

\end{itemize}

\bibliography{acl2020,anthology}

\begin{thebibliography}{78}
\expandafter\ifx\csname natexlab\endcsname\relax\def\natexlab#1{#1}\fi

\bibitem[{Adi et~al.(2016)Adi, Kermany, Belinkov, Lavi, and
  Goldberg}]{adi2016fine}
Yossi Adi, Einat Kermany, Yonatan Belinkov, Ofer Lavi, and Yoav Goldberg. 2016.
\newblock {Fine-grained Analysis of Sentence Embeddings Using Auxiliary
  Prediction Tasks}.
\newblock \emph{arXiv preprint arXiv:1608.04207}.

\bibitem[{Alammar(2020)}]{alammar2020explaining}
J~Alammar. 2020.
\newblock \href {https://jalammar.github.io/explaining-transformers/}
  {Interfaces for explaining transformer language models}.

\bibitem[{Alammar(2021)}]{alammar-2021-ecco}
J~Alammar. 2021.
\newblock \href {https://doi.org/10.18653/v1/2021.acl-demo.30} {Ecco: An open
  source library for the explainability of transformer language models}.
\newblock In \emph{Proceedings of the 59th Annual Meeting of the Association
  for Computational Linguistics and the 11th International Joint Conference on
  Natural Language Processing: System Demonstrations}, pages 249--257, Online.
  Association for Computational Linguistics.

\bibitem[{Amjad et~al.(2018)Amjad, Liu, and Geiger}]{ranaAblation}
Rana~Ali Amjad, Kairen Liu, and Bernhard~C. Geiger. 2018.
\newblock \href {https://doi.org/10.48550/ARXIV.1804.06679} {Understanding
  neural networks and individual neuron importance via information-ordered
  cumulative ablation}.

\bibitem[{Antverg and Belinkov(2022)}]{antverg2022on}
Omer Antverg and Yonatan Belinkov. 2022.
\newblock \href {https://openreview.net/forum?id=8uz0EWPQIMu} {On the pitfalls
  of analyzing individual neurons in language models}.
\newblock In \emph{International Conference on Learning Representations}.

\bibitem[{Antverg et~al.(2022)Antverg, Ben-David, and Belinkov}]{IDANI}
Omer Antverg, Eyal Ben-David, and Yonatan Belinkov. 2022.
\newblock \href {https://doi.org/10.48550/ARXIV.2206.00259} {Idani:
  Inference-time domain adaptation via neuron-level interventions}.

\bibitem[{Bahdanau et~al.(2014)Bahdanau, Cho, and Bengio}]{bahdanau2014neural}
Dzmitry Bahdanau, Kyunghyun Cho, and Yoshua Bengio. 2014.
\newblock {Neural Machine Translation by Jointly Learning to Align and
  Translate}.
\newblock \emph{arXiv preprint arXiv:1409.0473}.

\bibitem[{Bau et~al.(2019)Bau, Belinkov, Sajjad, Durrani, Dalvi, and
  Glass}]{bau2018identifying}
Anthony Bau, Yonatan Belinkov, Hassan Sajjad, Nadir Durrani, Fahim Dalvi, and
  James Glass. 2019.
\newblock \href {https://openreview.net/forum?id=H1z-PsR5KX} {Identifying and
  controlling important neurons in neural machine translation}.
\newblock In \emph{International Conference on Learning Representations}.

\bibitem[{Belinkov et~al.(2017{\natexlab{a}})Belinkov, Durrani, Dalvi, Sajjad,
  and Glass}]{belinkov:2017:acl}
Yonatan Belinkov, Nadir Durrani, Fahim Dalvi, Hassan Sajjad, and James Glass.
  2017{\natexlab{a}}.
\newblock \href
  {https://aclanthology.coli.uni-saarland.de/pdf/P/P17/P17-1080.pdf} {{What do
  Neural Machine Translation Models Learn about Morphology?}}
\newblock In \emph{Proceedings of the 55th Annual Meeting of the Association
  for Computational Linguistics (ACL)}, Vancouver. Association for
  Computational Linguistics.

\bibitem[{Belinkov et~al.(2020{\natexlab{a}})Belinkov, Durrani, Dalvi, Sajjad,
  and Glass}]{belinkov-etal-2020-analysis}
Yonatan Belinkov, Nadir Durrani, Fahim Dalvi, Hassan Sajjad, and James Glass.
  2020{\natexlab{a}}.
\newblock On the linguistic representational power of neural machine
  translation models.
\newblock \emph{Computational Linguistics}, 45(1):1--57.

\bibitem[{Belinkov et~al.(2020{\natexlab{b}})Belinkov, Durrani, Dalvi, Sajjad,
  and Glass}]{belinkov-etal-2020-linguistic}
Yonatan Belinkov, Nadir Durrani, Fahim Dalvi, Hassan Sajjad, and James Glass.
  2020{\natexlab{b}}.
\newblock \href {https://doi.org/10.1162/coli_a_00367} {On the linguistic
  representational power of neural machine translation models}.
\newblock \emph{Computational Linguistics}, 46(1):1--52.

\bibitem[{Belinkov et~al.(2017{\natexlab{b}})Belinkov, M{\`a}rquez, Sajjad,
  Durrani, Dalvi, and Glass}]{belinkov-etal-2017-evaluating}
Yonatan Belinkov, Llu{\'\i}s M{\`a}rquez, Hassan Sajjad, Nadir Durrani, Fahim
  Dalvi, and James Glass. 2017{\natexlab{b}}.
\newblock \href {https://www.aclweb.org/anthology/I17-1001} {Evaluating layers
  of representation in neural machine translation on part-of-speech and
  semantic tagging tasks}.
\newblock In \emph{Proceedings of the Eighth International Joint Conference on
  Natural Language Processing (Volume 1: Long Papers)}, pages 1--10, Taipei,
  Taiwan. Asian Federation of Natural Language Processing.

\bibitem[{Bengio et~al.(2012)Bengio, Courville, and
  Vincent}]{bengio_disentangled}
Yoshua Bengio, Aaron~C. Courville, and Pascal Vincent. 2012.
\newblock \href {http://arxiv.org/abs/1206.5538} {Unsupervised feature learning
  and deep learning: {A} review and new perspectives}.
\newblock \emph{CoRR}, abs/1206.5538.

\bibitem[{Binshtok et~al.(2007)Binshtok, Brafman, Shimony, Martin, and
  Boutilier}]{binshtok:2007:AAAI}
Maxim Binshtok, Ronen~I Brafman, Solomon~Eyal Shimony, Ajay Martin, and Crag
  Boutilier. 2007.
\newblock Computing optimal subsets.
\newblock In \emph{Proceedings of the Twenty Second AAAI Conference on
  Artificial Intelligence (AAAI, Oral presentation)}.

\bibitem[{Conneau et~al.(2018)Conneau, Kruszewski, Lample, Barrault, and
  Baroni}]{conneau2018you}
Alexis Conneau, German Kruszewski, Guillaume Lample, Lo{\"\i}c Barrault, and
  Marco Baroni. 2018.
\newblock {What you can cram into a single vector: Probing sentence embeddings
  for linguistic properties}.
\newblock In \emph{Proceedings of the 56th Annual Meeting of the Association
  for Computational Linguistics (ACL)}.

\bibitem[{Dai et~al.(2021)Dai, Dong, Hao, Sui, and Wei}]{knowledgeneurons}
Damai Dai, Li~Dong, Yaru Hao, Zhifang Sui, and Furu Wei. 2021.
\newblock Knowledge neurons in pretrained transformers.
\newblock \emph{CoRR}, abs/2104.08696.

\bibitem[{Dalvi et~al.(2019)Dalvi, Durrani, Sajjad, Belinkov, Bau, and
  Glass}]{dalvi:2019:AAAI}
Fahim Dalvi, Nadir Durrani, Hassan Sajjad, Yonatan Belinkov, D.~Anthony Bau,
  and James Glass. 2019.
\newblock What is one grain of sand in the desert? analyzing individual neurons
  in deep nlp models.
\newblock In \emph{Proceedings of the Thirty-Third AAAI Conference on
  Artificial Intelligence (AAAI, Oral presentation)}.

\bibitem[{Dalvi et~al.(2022)Dalvi, Khan, Alam, Durrani, Xu, and
  Sajjad}]{dalvi2022discovering}
Fahim Dalvi, Abdul~Rafae Khan, Firoj Alam, Nadir Durrani, Jia Xu, and Hassan
  Sajjad. 2022.
\newblock \href {https://openreview.net/forum?id=POTMtpYI1xH} {Discovering
  latent concepts learned in {BERT}}.
\newblock In \emph{International Conference on Learning Representations}.

\bibitem[{Dalvi et~al.(2020)Dalvi, Sajjad, Durrani, and
  Belinkov}]{dalvi-2020-CCFS}
Fahim Dalvi, Hassan Sajjad, Nadir Durrani, and Yonatan Belinkov. 2020.
\newblock Analyzing redundancy in pretrained transformer models.
\newblock In \emph{Proceedings of the 2020 Conference on Empirical Methods in
  Natural Language Processing (EMNLP-2020)}, Online.

\bibitem[{De~Cao et~al.(2020)De~Cao, Schlichtkrull, Aziz, and
  Titov}]{de-cao-etal-2020-decisions}
Nicola De~Cao, Michael~Sejr Schlichtkrull, Wilker Aziz, and Ivan Titov. 2020.
\newblock \href {https://doi.org/10.18653/v1/2020.emnlp-main.262} {How do
  decisions emerge across layers in neural models? interpretation with
  differentiable masking}.
\newblock In \emph{Proceedings of the 2020 Conference on Empirical Methods in
  Natural Language Processing (EMNLP)}, pages 3243--3255, Online. Association
  for Computational Linguistics.

\bibitem[{Devlin et~al.(2019)Devlin, Chang, Lee, and
  Toutanova}]{devlin-etal-2019-bert}
Jacob Devlin, Ming-Wei Chang, Kenton Lee, and Kristina Toutanova. 2019.
\newblock {BERT}: Pre-training of deep bidirectional transformers for language
  understanding.
\newblock In \emph{Proceedings of the 2019 Conference of the North {A}merican
  Chapter of the Association for Computational Linguistics: Human Language
  Technologies, Volume 1 (Long and Short Papers)}, Minneapolis, Minnesota.
  Association for Computational Linguistics.

\bibitem[{Dhamdhere et~al.(2020)Dhamdhere, Agarwal, and
  Sundararajan}]{dhamdhere2020shapley}
Kedar Dhamdhere, Ashish Agarwal, and Mukund Sundararajan. 2020.
\newblock \href {http://arxiv.org/abs/1902.05622} {The shapley taylor
  interaction index}.

\bibitem[{Dhamdhere et~al.(2018)Dhamdhere, Sundararajan, and
  Yan}]{DBLP:journals/corr/abs-1805-12233}
Kedar Dhamdhere, Mukund Sundararajan, and Qiqi Yan. 2018.
\newblock \href {http://arxiv.org/abs/1805.12233} {How important is a neuron?}
\newblock \emph{CoRR}, abs/1805.12233.

\bibitem[{Durrani et~al.(2022)Durrani, Dalvi, and Sajjad}]{LCAJournal}
Nadir Durrani, Fahim Dalvi, and Hassan Sajjad. 2022.
\newblock \href {https://doi.org/10.48550/ARXIV.2206.13288} {Linguistic
  correlation analysis: Discovering salient neurons in deepnlp models}.

\bibitem[{Durrani et~al.(2021)Durrani, Sajjad, and Dalvi}]{durrani-FT}
Nadir Durrani, Hassan Sajjad, and Fahim Dalvi. 2021.
\newblock How transfer learning impacts linguistic knowledge in deep nlp
  models?
\newblock In \emph{Findings of the Association for Computational Linguistics:
  ACL 2021}, Online. Association for Computational Linguistics.

\bibitem[{Durrani et~al.(2020)Durrani, Sajjad, Dalvi, and
  Belinkov}]{durrani-etal-2020-analyzing}
Nadir Durrani, Hassan Sajjad, Fahim Dalvi, and Yonatan Belinkov. 2020.
\newblock \href {https://doi.org/10.18653/v1/2020.emnlp-main.395} {Analyzing
  individual neurons in pre-trained language models}.
\newblock In \emph{Proceedings of the 2020 Conference on Empirical Methods in
  Natural Language Processing (EMNLP)}, pages 4865--4880, Online. Association
  for Computational Linguistics.

\bibitem[{Elazar et~al.(2021)Elazar, Ravfogel, Jacovi, and
  Goldberg}]{elazar-etal-2021-amnesic}
Yanai Elazar, Shauli Ravfogel, Alon Jacovi, and Yoav Goldberg. 2021.
\newblock \href {https://doi.org/10.1162/tacl_a_00359} {Amnesic probing:
  Behavioral explanation with amnesic counterfactuals}.
\newblock \emph{Transactions of the Association for Computational Linguistics},
  9:160--175.

\bibitem[{Erhan et~al.(2009)Erhan, Bengio, Courville, and
  Vincent}]{erhan_gradient_ascent_techreport}
Dumitru Erhan, Yoshua Bengio, Aaron Courville, and Pascal Vincent. 2009.
\newblock Visualizing higher-layer features of a deep network.
\newblock Technical Report 1341, University of Montreal.
\newblock Also presented at the ICML 2009 Workshop on Learning Feature
  Hierarchies, Montr{'{e}}al, Canada.

\bibitem[{Faruqui et~al.(2015)Faruqui, Tsvetkov, Yogatama, Dyer, and
  Smith}]{faruqui-etal-2015-sparse}
Manaal Faruqui, Yulia Tsvetkov, Dani Yogatama, Chris Dyer, and Noah~A. Smith.
  2015.
\newblock \href {https://doi.org/10.3115/v1/P15-1144} {Sparse overcomplete word
  vector representations}.
\newblock In \emph{Proceedings of the 53rd Annual Meeting of the Association
  for Computational Linguistics and the 7th International Joint Conference on
  Natural Language Processing (Volume 1: Long Papers)}, pages 1491--1500,
  Beijing, China. Association for Computational Linguistics.

\bibitem[{Feder et~al.(2021)Feder, Oved, Shalit, and
  Reichart}]{feder-etal-2021-causalm}
Amir Feder, Nadav Oved, Uri Shalit, and Roi Reichart. 2021.
\newblock \href {https://doi.org/10.1162/coli_a_00404} {{C}ausa{LM}: Causal
  model explanation through counterfactual language models}.
\newblock \emph{Computational Linguistics}, 47(2):333--386.

\bibitem[{Feng et~al.(2018)Feng, Wallace, Grissom~II, Iyyer, Rodriguez, and
  Boyd-Graber}]{feng-etal-2018-pathologies}
Shi Feng, Eric Wallace, Alvin Grissom~II, Mohit Iyyer, Pedro Rodriguez, and
  Jordan Boyd-Graber. 2018.
\newblock \href {https://doi.org/10.18653/v1/D18-1407} {Pathologies of neural
  models make interpretations difficult}.
\newblock In \emph{Proceedings of the 2018 Conference on Empirical Methods in
  Natural Language Processing}, pages 3719--3728, Brussels, Belgium.
  Association for Computational Linguistics.

\bibitem[{Frankle and Carbin(2019)}]{frankle2018_lotterticket}
Jonathan Frankle and Michael Carbin. 2019.
\newblock \href {https://openreview.net/forum?id=rJl-b3RcF7} {The lottery
  ticket hypothesis: Finding sparse, trainable neural networks}.
\newblock In \emph{International Conference on Learning Representations}.

\bibitem[{Fyshe et~al.(2015)Fyshe, Wehbe, Talukdar, Murphy, and
  Mitchell}]{fyshe-etal-2015-compositional}
Alona Fyshe, Leila Wehbe, Partha~P. Talukdar, Brian Murphy, and Tom~M.
  Mitchell. 2015.
\newblock \href {https://doi.org/10.3115/v1/N15-1004} {A compositional and
  interpretable semantic space}.
\newblock In \emph{Proceedings of the 2015 Conference of the North {A}merican
  Chapter of the Association for Computational Linguistics: Human Language
  Technologies}, pages 32--41, Denver, Colorado. Association for Computational
  Linguistics.

\bibitem[{Godin et~al.(2018)Godin, Demuynck, Dambre, De~Neve, and
  Demeester}]{godin-etal-2018-explaining}
Fr{\'e}deric Godin, Kris Demuynck, Joni Dambre, Wesley De~Neve, and Thomas
  Demeester. 2018.
\newblock \href {https://doi.org/10.18653/v1/D18-1365} {Explaining
  character-aware neural networks for word-level prediction: Do they discover
  linguistic rules?}
\newblock In \emph{Proceedings of the 2018 Conference on Empirical Methods in
  Natural Language Processing}, pages 3275--3284, Brussels, Belgium.
  Association for Computational Linguistics.

\bibitem[{Gu et~al.(2021)Gu, Feng, and Xie}]{gu2021pruningthenexpanding}
Shuhao Gu, Yang Feng, and Wanying Xie. 2021.
\newblock \href {http://arxiv.org/abs/2103.13678} {Pruning-then-expanding model
  for domain adaptation of neural machine translation}.

\bibitem[{Hennigen et~al.(2020)Hennigen, Williams, and
  Cotterell}]{torroba-hennigen-etal-2020-intrinsic}
Lucas~Torroba Hennigen, Adina Williams, and Ryan Cotterell. 2020.
\newblock \href {https://doi.org/10.18653/v1/2020.emnlp-main.15} {Intrinsic
  probing through dimension selection}.
\newblock In \emph{Proceedings of the 2020 Conference on Empirical Methods in
  Natural Language Processing (EMNLP)}, pages 197--216, Online. Association for
  Computational Linguistics.

\bibitem[{Hewitt and Liang(2019)}]{hewitt-liang-2019-designing}
John Hewitt and Percy Liang. 2019.
\newblock \href {https://doi.org/10.18653/v1/D19-1275} {Designing and
  interpreting probes with control tasks}.
\newblock In \emph{Proceedings of the 2019 Conference on Empirical Methods in
  Natural Language Processing and the 9th International Joint Conference on
  Natural Language Processing (EMNLP-IJCNLP)}, pages 2733--2743, Hong Kong,
  China.

\bibitem[{Hupkes(2020)}]{hupkes2020hierarchy}
Dieuwke Hupkes. 2020.
\newblock \emph{Hierarchy and interpretability in neural models of language
  processing}.
\newblock Ph.D. thesis, University of Amsterdam.

\bibitem[{Hupkes et~al.(2018)Hupkes, Veldhoen, and
  Zuidema}]{hupkes2018visualisation}
Dieuwke Hupkes, Sara Veldhoen, and Willem Zuidema. 2018.
\newblock \href {http://arxiv.org/abs/1711.10203} {Visualisation and
  'diagnostic classifiers' reveal how recurrent and recursive neural networks
  process hierarchical structure}.

\bibitem[{Jones et~al.(2012)Jones, Andreas, Bauer, Hermann, and
  Knight}]{jones-etal-2012-semantics}
Bevan Jones, Jacob Andreas, Daniel Bauer, Karl~Moritz Hermann, and Kevin
  Knight. 2012.
\newblock \href {https://aclanthology.org/C12-1083} {Semantics-based machine
  translation with hyperedge replacement grammars}.
\newblock In \emph{Proceedings of {COLING} 2012}, pages 1359--1376, Mumbai,
  India. The COLING 2012 Organizing Committee.

\bibitem[{K{\'a}d{\'a}r et~al.(2017)K{\'a}d{\'a}r, Chrupa{\l}a, and
  Alishahi}]{kadar-etal-2017-representation}
{\'A}kos K{\'a}d{\'a}r, Grzegorz Chrupa{\l}a, and Afra Alishahi. 2017.
\newblock \href {https://doi.org/10.1162/COLI_a_00300} {Representation of
  linguistic form and function in recurrent neural networks}.
\newblock \emph{Computational Linguistics}, 43(4):761--780.

\bibitem[{Karpathy et~al.(2015)Karpathy, Johnson, and
  Fei-Fei}]{karpathy2015visualizing}
Andrej Karpathy, Justin Johnson, and Li~Fei-Fei. 2015.
\newblock Visualizing and understanding recurrent networks.
\newblock \emph{arXiv preprint arXiv:1506.02078}.

\bibitem[{Lakretz et~al.(2019)Lakretz, Kruszewski, Desbordes, Hupkes, Dehaene,
  and Baroni}]{lakretz-etal-2019-emergence}
Yair Lakretz, German Kruszewski, Theo Desbordes, Dieuwke Hupkes, Stanislas
  Dehaene, and Marco Baroni. 2019.
\newblock \href {https://doi.org/10.18653/v1/N19-1002} {The emergence of number
  and syntax units in {LSTM} language models}.
\newblock In \emph{Proceedings of the 2019 Conference of the North {A}merican
  Chapter of the Association for Computational Linguistics: Human Language
  Technologies, Volume 1 (Long and Short Papers)}, pages 11--20, Minneapolis,
  Minnesota. Association for Computational Linguistics.

\bibitem[{Li et~al.(2016{\natexlab{a}})Li, Chen, Hovy, and
  Jurafsky}]{li-etal-2016-visualizing}
Jiwei Li, Xinlei Chen, Eduard Hovy, and Dan Jurafsky. 2016{\natexlab{a}}.
\newblock \href {https://doi.org/10.18653/v1/N16-1082} {Visualizing and
  understanding neural models in {NLP}}.
\newblock In \emph{Proceedings of the 2016 Conference of the North {A}merican
  Chapter of the Association for Computational Linguistics: Human Language
  Technologies}, pages 681--691, San Diego, California. Association for
  Computational Linguistics.

\bibitem[{Li et~al.(2016{\natexlab{b}})Li, Monroe, and Jurafsky}]{LiMJ16a}
Jiwei Li, Will Monroe, and Dan Jurafsky. 2016{\natexlab{b}}.
\newblock \href {http://arxiv.org/abs/1612.08220} {Understanding neural
  networks through representation erasure}.
\newblock \emph{CoRR}, abs/1612.08220.

\bibitem[{Lipton(2016)}]{lipton2016mythos}
Zachary~C Lipton. 2016.
\newblock {The Mythos of Model Interpretability}.
\newblock In \emph{ICML Workshop on Human Interpretability in Machine Learning
  (WHI)}.

\bibitem[{Liu et~al.(2019)Liu, Gardner, Belinkov, Peters, and
  Smith}]{liu-etal-2019-linguistic}
Nelson~F. Liu, Matt Gardner, Yonatan Belinkov, Matthew~E. Peters, and Noah~A.
  Smith. 2019.
\newblock \href {https://www.aclweb.org/anthology/N19-1112} {Linguistic
  knowledge and transferability of contextual representations}.
\newblock In \emph{Proceedings of the 2019 Conference of the North {A}merican
  Chapter of the Association for Computational Linguistics: Human Language
  Technologies, Volume 1 (Long and Short Papers)}, pages 1073--1094,
  Minneapolis, Minnesota. Association for Computational Linguistics.

\bibitem[{Lundberg and Lee(2017)}]{shappely_NIPS2017_7062}
Scott~M Lundberg and Su-In Lee. 2017.
\newblock \href
  {http://papers.nips.cc/paper/7062-a-unified-approach-to-interpreting-model-predictions.pdf}
  {A unified approach to interpreting model predictions}.
\newblock In I.~Guyon, U.~V. Luxburg, S.~Bengio, H.~Wallach, R.~Fergus,
  S.~Vishwanathan, and R.~Garnett, editors, \emph{Advances in Neural
  Information Processing Systems 30}, pages 4765--4774. Curran Associates, Inc.

\bibitem[{Marasović(2018)}]{MarasovicGradient2018NLP}
Ana Marasović. 2018.
\newblock Nlp’s generalization problem, and how researchers are tackling it.
\newblock \emph{The Gradient}.

\bibitem[{Mclnnes et~al.(2020)Mclnnes, Healy, and Melville}]{mcinnes2020umap}
Leland Mclnnes, John Healy, and James Melville. 2020.
\newblock \href {http://arxiv.org/abs/1802.03426} {{UMAP}: Uniform manifold
  approximation and projection for dimension reduction}.

\bibitem[{Meyes et~al.(2020)Meyes, de~Puiseau, Posada{-}Moreno, and
  Meisen}]{mayes_under_hood:2020}
Richard Meyes, Constantin~Waubert de~Puiseau, Andres Posada{-}Moreno, and
  Tobias Meisen. 2020.
\newblock Under the hood of neural networks: Characterizing learned
  representations by functional neuron populations and network ablations.
\newblock \emph{CoRR}, abs/2004.01254.

\bibitem[{Michael et~al.(2020)Michael, Botha, and
  Tenney}]{michael-etal-2020-asking}
Julian Michael, Jan~A. Botha, and Ian Tenney. 2020.
\newblock \href {https://doi.org/10.18653/v1/2020.emnlp-main.552} {Asking
  without telling: Exploring latent ontologies in contextual representations}.
\newblock In \emph{Proceedings of the 2020 Conference on Empirical Methods in
  Natural Language Processing (EMNLP)}, pages 6792--6812, Online. Association
  for Computational Linguistics.

\bibitem[{Mikolov et~al.(2013)Mikolov, Chen, Corrado, and
  Dean}]{Mikolov:2013:ICLR}
Tomas Mikolov, Kai Chen, Greg Corrado, and Jeffrey Dean. 2013.
\newblock Efficient estimation of word representations in vector space.
\newblock In \emph{Proceedings of the ICLR Workshop}, Scottsdale, AZ, USA.

\bibitem[{Mu and Andreas(2020)}]{Mu-Nips}
Jesse Mu and Jacob Andreas. 2020.
\newblock \href {http://arxiv.org/abs/2006.14032} {Compositional explanations
  of neurons}.
\newblock \emph{CoRR}, abs/2006.14032.

\bibitem[{Murdoch et~al.(2018)Murdoch, Liu, and Yu}]{contextualDecomposition}
W.~James Murdoch, Peter~J. Liu, and Bin Yu. 2018.
\newblock \href {https://doi.org/10.48550/ARXIV.1801.05453} {Beyond word
  importance: Contextual decomposition to extract interactions from lstms}.

\bibitem[{Na et~al.(2019)Na, Choe, Lee, and Kim}]{Na-ICLR}
Seil Na, Yo~Joong Choe, Dong{-}Hyun Lee, and Gunhee Kim. 2019.
\newblock \href {http://arxiv.org/abs/1902.07249} {Discovery of natural
  language concepts in individual units of {CNNs}}.
\newblock \emph{CoRR}, abs/1902.07249.

\bibitem[{Olah et~al.(2018)Olah, Satyanarayan, Johnson, Carter, Schubert, Ye,
  and Mordvintsev}]{olah2018the}
Chris Olah, Arvind Satyanarayan, Ian Johnson, Shan Carter, Ludwig Schubert,
  Katherine Ye, and Alexander Mordvintsev. 2018.
\newblock \href {https://doi.org/10.23915/distill.00010} {The building blocks
  of interpretability}.
\newblock \emph{Distill}.
\newblock Https://distill.pub/2018/building-blocks.

\bibitem[{Pimentel et~al.(2020)Pimentel, Valvoda, Hall~Maudslay, Zmigrod,
  Williams, and Cotterell}]{pimentel2020informationtheoretic}
Tiago Pimentel, Josef Valvoda, Rowan Hall~Maudslay, Ran Zmigrod, Adina
  Williams, and Ryan Cotterell. 2020.
\newblock \href {https://doi.org/10.18653/v1/2020.acl-main.420}
  {Information-theoretic probing for linguistic structure}.
\newblock In \emph{Proceedings of the 58th Annual Meeting of the Association
  for Computational Linguistics}, pages 4609--4622, Online. Association for
  Computational Linguistics.

\bibitem[{Poerner et~al.(2018)Poerner, Roth, and
  Sch{\"u}tze}]{poerner-etal-2018-interpretable}
Nina Poerner, Benjamin Roth, and Hinrich Sch{\"u}tze. 2018.
\newblock \href {https://doi.org/10.18653/v1/W18-5437} {Interpretable textual
  neuron representations for {NLP}}.
\newblock In \emph{Proceedings of the 2018 {EMNLP} Workshop {B}lackbox{NLP}:
  Analyzing and Interpreting Neural Networks for {NLP}}, pages 325--327,
  Brussels, Belgium. Association for Computational Linguistics.

\bibitem[{Radford et~al.(2019)Radford, Wu, Child, Luan, Amodei, Sutskever
  et~al.}]{Radford}
Alec Radford, Jeffrey Wu, Rewon Child, David Luan, Dario Amodei, Ilya
  Sutskever, et~al. 2019.
\newblock Language models are unsupervised multitask learners.
\newblock \emph{OpenAI blog}, 1(8):9.

\bibitem[{Raghu et~al.(2017)Raghu, Gilmer, Yosinski, and
  Sohl-Dickstein}]{NIPS2017_7188_svcca}
Maithra Raghu, Justin Gilmer, Jason Yosinski, and Jascha Sohl-Dickstein. 2017.
\newblock \href
  {http://papers.nips.cc/paper/7188-svcca-singular-vector-canonical-correlation-analysis-for-deep-learning-dynamics-and-interpretability.pdf}
  {{SVCCA: Singular Vector Canonical Correlation Analysis for Deep Learning
  Dynamics and Interpretability}}.
\newblock In I.~Guyon, U.~V. Luxburg, S.~Bengio, H.~Wallach, R.~Fergus,
  S.~Vishwanathan, and R.~Garnett, editors, \emph{Advances in Neural
  Information Processing Systems 30}, pages 6078--6087. Curran Associates, Inc.

\bibitem[{Sajjad et~al.(2021)Sajjad, Alam, Dalvi, and
  Durrani}]{sajjad2021:znorm}
Hassan Sajjad, Firoj Alam, Fahim Dalvi, and Nadir Durrani. 2021.
\newblock \href {http://arxiv.org/abs/2104.07456} {Effect of post-processing on
  contextualized word representations}.
\newblock \emph{CoRR}, abs/2104.07456.

\bibitem[{Sajjad et~al.(2022)Sajjad, Durrani, Dalvi, Alam, Khan, and
  Xu}]{sajjad-etal-2022-analyzing}
Hassan Sajjad, Nadir Durrani, Fahim Dalvi, Firoj Alam, Abdul Khan, and Jia Xu.
  2022.
\newblock \href {https://doi.org/10.18653/v1/2022.naacl-main.225} {Analyzing
  encoded concepts in transformer language models}.
\newblock In \emph{Proceedings of the 2022 Conference of the North American
  Chapter of the Association for Computational Linguistics: Human Language
  Technologies}, pages 3082--3101, Seattle, United States. Association for
  Computational Linguistics.

\bibitem[{Seyffarth et~al.(2021)Seyffarth, Samih, Kallmeyer, and
  Sajjad}]{esther_causativity_iwcs21}
Esther Seyffarth, Younes Samih, Laura Kallmeyer, and Hassan Sajjad. 2021.
\newblock Implicit representations of event properties within contextual
  language models: Searching for ``causativity neurons".
\newblock In \emph{International Conference on Computational Semantics (IWCS)}.

\bibitem[{Stanczak et~al.(2022)Stanczak, Hennigen, Williams, Cotterell, and
  Augenstein}]{intrinsicProbing2}
Karolina Stanczak, Lucas~Torroba Hennigen, Adina Williams, Ryan Cotterell, and
  Isabelle Augenstein. 2022.
\newblock \href {http://arxiv.org/abs/2201.08214} {A latent-variable model for
  intrinsic probing}.
\newblock \emph{CoRR}, abs/2201.08214.

\bibitem[{Stock(2010)}]{stock2010concepts}
Wolfgang~G Stock. 2010.
\newblock Concepts and semantic relations in information science.
\newblock \emph{Journal of the American Society for Information Science and
  Technology}, 61(10):1951--1969.

\bibitem[{Suau et~al.(2020)Suau, Zappella, and Apostoloff}]{suau2020finding}
Xavier Suau, Luca Zappella, and Nicholas Apostoloff. 2020.
\newblock \href {http://arxiv.org/abs/2005.07647} {Finding experts in
  transformer models}.
\newblock \emph{CoRR}, abs/2005.07647.

\bibitem[{Sundararajan et~al.(2017)Sundararajan, Taly, and Yan}]{ig}
Mukund Sundararajan, Ankur Taly, and Qiqi Yan. 2017.
\newblock \href {http://arxiv.org/abs/1703.01365} {Axiomatic attribution for
  deep networks}.

\bibitem[{Sutskever et~al.(2014)Sutskever, Vinyals, and Le}]{SutskeverVL14}
Ilya Sutskever, Oriol Vinyals, and Quoc~V. Le. 2014.
\newblock \href {http://arxiv.org/abs/1409.3215} {Sequence to sequence learning
  with neural networks}.
\newblock \emph{CoRR}, abs/1409.3215.

\bibitem[{Tenney et~al.(2019)Tenney, Das, and Pavlick}]{tenney-etal-2019-bert}
Ian Tenney, Dipanjan Das, and Ellie Pavlick. 2019.
\newblock \href {https://doi.org/10.18653/v1/P19-1452} {{BERT} rediscovers the
  classical {NLP} pipeline}.
\newblock In \emph{Proceedings of the 57th Annual Meeting of the Association
  for Computational Linguistics}, pages 4593--4601, Florence, Italy.
  Association for Computational Linguistics.

\bibitem[{Tran et~al.(2018)Tran, Bisazza, and Monz}]{tran2018importance}
Ke~Tran, Arianna Bisazza, and Christof Monz. 2018.
\newblock {The Importance of Being Recurrent for Modeling Hierarchical
  Structure}.
\newblock \emph{arXiv preprint arXiv:1803.03585}.

\bibitem[{Valipour et~al.(2019)Valipour, Lee, Jamacaro, and
  Bessega}]{valipur-2019}
Mehrdad Valipour, En{-}Shiun~Annie Lee, Jaime~R. Jamacaro, and Carolina
  Bessega. 2019.
\newblock \href {http://arxiv.org/abs/1912.05308} {Unsupervised transfer
  learning via {BERT} neuron selection}.
\newblock \emph{CoRR}, abs/1912.05308.

\bibitem[{Vauquois(1968)}]{Vauquois68}
Bernard Vauquois. 1968.
\newblock A survey of formal grammars and algorithms for recognition and
  transformation in mechanical translation.
\newblock In \emph{IFIP Congress (2)}, pages 1114--1122.

\bibitem[{Voita and Titov(2020)}]{voita2020informationtheoretic}
Elena Voita and Ivan Titov. 2020.
\newblock Information-theoretic probing with minimum description length.
\newblock In \emph{Proceedings of the 2020 Conference on Empirical Methods in
  Natural Language Processing}. Association for Computational Linguistics.

\bibitem[{Wu et~al.(2020)Wu, Belinkov, Sajjad, Durrani, Dalvi, and
  Glass}]{wu:2020:acl}
John Wu, Yonatan Belinkov, Hassan Sajjad, Nadir Durrani, Fahim Dalvi, and James
  Glass. 2020.
\newblock {Similarity Analysis of Contextual Word Representation Models}.
\newblock In \emph{Proceedings of the 58th Annual Meeting of the Association
  for Computational Linguistics (ACL)}, Seattle. Association for Computational
  Linguistics.

\bibitem[{Xin et~al.(2019)Xin, Lin, and Yu}]{xin-etal-2019-part}
Ji~Xin, Jimmy Lin, and Yaoliang Yu. 2019.
\newblock \href {https://doi.org/10.18653/v1/D19-1591} {What part of the neural
  network does this? understanding {LSTM}s by measuring and dissecting
  neurons}.
\newblock In \emph{Proceedings of the 2019 Conference on Empirical Methods in
  Natural Language Processing and the 9th International Joint Conference on
  Natural Language Processing (EMNLP-IJCNLP)}, pages 5823--5830, Hong Kong,
  China. Association for Computational Linguistics.

\bibitem[{Zhang and Bowman(2018)}]{zhang-bowman-2018-language}
Kelly Zhang and Samuel Bowman. 2018.
\newblock \href {https://doi.org/10.18653/v1/W18-5448} {Language modeling
  teaches you more than translation does: Lessons learned through auxiliary
  syntactic task analysis}.
\newblock In \emph{Proceedings of the 2018 {EMNLP} Workshop {B}lackbox{NLP}:
  Analyzing and Interpreting Neural Networks for {NLP}}, pages 359--361,
  Brussels, Belgium. Association for Computational Linguistics.

\bibitem[{Zintgraf et~al.(2017)Zintgraf, Cohen, Adel, and
  Welling}]{ZintgrafCAW17}
Luisa~M. Zintgraf, Taco~S. Cohen, Tameem Adel, and Max Welling. 2017.
\newblock \href {http://arxiv.org/abs/1702.04595} {Visualizing deep neural
  network decisions: Prediction difference analysis}.
\newblock \emph{CoRR}, abs/1702.04595.

\end{thebibliography}
\bibliographystyle{acl_natbib}

\appendix
\newpage

\begin{table*}[t]
\centering
\footnotesize
\begin{tabular}{p{0.19\textwidth}p{0.08\textwidth}p{0.1\textwidth}p{0.1\textwidth}p{0.1\textwidth}p{0.06\textwidth}p{0.1\textwidth}p{0.06\textwidth}}
\toprule
   & Scope & Input & Output  & Scalability & HITL & Supervision & Causation \\
\midrule
\rowcolor[HTML]{D9EAD3} 
\multicolumn{8}{l}{\cellcolor[HTML]{D9EAD3}\textbf{Visualization}}                                                                                                                        \\
\rowcolor[HTML]{D9EAD3} 
\citet{karpathy2015visualizing}  & local        & neuron                 & concept              & low         & yes  & no              & no            \\
\rowcolor[HTML]{D9EAD3} 
\citet{fyshe-etal-2015-compositional, faruqui-etal-2015-sparse, li-etal-2016-visualizing}  &         &                  &               &           &    &                &             \\
\rowcolor[HTML]{A4C2F4} 
\multicolumn{8}{l}{\cellcolor[HTML]{A4C2F4}\textbf{Corpus-based methods}}                                                              
                                     \\
\multicolumn{8}{l}{\cellcolor[HTML]{A4C2F4}{Concept Search}}                                                                                                                  \\
\rowcolor[HTML]{A4C2F4} 
\cellcolor[HTML]{A4C2F4}    \citet{kadar-etal-2017-representation} & global       & neuron                 & concept              & low         & yes  & no              & no            \\
\rowcolor[HTML]{A4C2F4} \citet{Na-ICLR} 
  & global       & neuron                 & concept              & high        & no   & no              & no            \\
\rowcolor[HTML]{A4C2F4} 
\multicolumn{8}{l}{\cellcolor[HTML]{A4C2F4}{Neuron Search}}                                                                                                                  \\
\rowcolor[HTML]{A4C2F4} 
\cellcolor[HTML]{A4C2F4}    \citet{Mu-Nips,suau2020finding,antverg2022on,IDANI}  & global       & concept                & neurons              & high        & no   & yes             & no            \\
\rowcolor[HTML]{EA9999} 
\multicolumn{8}{l}{\cellcolor[HTML]{EA9999}\textbf{Probing-based methods}}                                                                                                            \\
\rowcolor[HTML]{EA9999} Linear \cite{Radford,dalvi:2019:AAAI,lakretz-etal-2019-emergence, durrani-etal-2020-analyzing,durrani-FT,LCAJournal,hupkes2020hierarchy,IDANI}     & global       & concept                & neurons              & high        & no   & yes             & no
\\
\rowcolor[HTML]{EA9999} 
Random Forest    \cite{valipur-2019}  & global       & concept                & neurons              & high        & no   & yes             & no            \\
\rowcolor[HTML]{EA9999} 
Gaussian    \cite{torroba-hennigen-etal-2020-intrinsic,intrinsicProbing2}  & global       & concept                & neurons              & high        & no   & yes             & no            \\
\rowcolor[HTML]{D9D9D9} 
\multicolumn{8}{l}{\cellcolor[HTML]{D9D9D9}\textbf{Causation-based methods}}                                                                                                              \\
\rowcolor[HTML]{D9D9D9} 
Ablation    \cite{li-etal-2016-visualizing,ranaAblation,xin-etal-2019-part,lakretz-etal-2019-emergence}   & both         & concept/ class         & neurons              & medium      & no   & no              & yes           \\
\rowcolor[HTML]{D9D9D9} 
Knowledge attribution  \cite{DBLP:journals/corr/abs-1805-12233,dhamdhere2020shapley, shappely_NIPS2017_7062, tran2018importance, knowledgeneurons,contextualDecomposition,godin-etal-2018-explaining}   & local        & concept/ class         & neurons              & high        & no   & no              & yes           \\
\rowcolor[HTML]{FFE599} 
\multicolumn{8}{l}{\cellcolor[HTML]{FFE599}\textbf{Miscellaneous methods}}                                                                                                                \\
\rowcolor[HTML]{FFE599} 
Corpus generation \cite{poerner-etal-2018-interpretable}   & global       & neuron                 & concept              & low         & yes  & no              & no            \\
\rowcolor[HTML]{FFE599} 
Matrix factorization \cite{alammar2020explaining,alammar-2021-ecco}   & local        & neurons                      & neurons      & low         & yes  & no              & no            \\
\rowcolor[HTML]{FFE599} 
Clustering \cite{dalvi-2020-CCFS}      & global        & neurons                      & neurons      & high         & yes  & no    & no            \\
\rowcolor[HTML]{FFE599} 
Multi model search  \cite{bau2018identifying,wu:2020:acl} & global       & neurons                & neurons              & high        & yes  & no              & no \\
\bottomrule
\end{tabular}

\caption{Comparison of neuron analysis methods based on various attributes. The exhaustive list of citations for each method are provided in the text.} 
\label{tab:appendix:neuronmethods}
\end{table*}

\end{document}